\documentclass[10pt, a4paper]{article}

\usepackage{template/lrec-coling2024} 
\usepackage{soul}
\usepackage{subcaption}
\usepackage{float}

\newcommand\blfootnote[1]{%
  \begingroup
  \renewcommand\thefootnote{}\footnote{#1}%
  \addtocounter{footnote}{-1}%
  \endgroup
}

\title{Revisiting The Classics: A Study on Identifying and Rectifying Gender Stereotypes in Rhymes and Poems}

\name{Aditya Narayan Sankaran$^{\ast \dagger}$, Vigneshwaran Shankaran$^{\ast \mathsection}$ \\
\bf{ \large Sampath Lonka$^{\ddagger}$, Rajesh Sharma$^{\| \bot}$}}

\address{$^\dagger$Sri Sathya Sai Institute of Higher Learning,\\ $^\mathsection$GESIS - Leibniz Institute for the Social Sciences, \\$^\ddagger$Citridot Solutions, $^\|$University of Tartu, $^\bot$ CARDS, IIT Ropar\\
\{adityanarayans@sssihl.edu.in, vigneshwaran.shankaran@gesis.org,\\ 
lsampath@citridot.com, rajesh.sharma@ut.ee\}}

\abstract{
Rhymes and poems are a powerful medium for transmitting cultural norms and societal roles. However, the pervasive existence of gender stereotypes in these works perpetuates biased perceptions and limits the scope of individuals' identities. Past works have shown that stereotyping and prejudice emerge in early childhood, and developmental research on causal mechanisms is critical for understanding and controlling stereotyping and prejudice. This work contributes by gathering a dataset of rhymes and poems to identify gender stereotypes and propose a model with 97\% accuracy to identify gender bias. Gender stereotypes were rectified using a Large Language Model (LLM) and its effectiveness was evaluated in a comparative survey against human educator rectifications. To summarize, this work highlights the pervasive nature of gender stereotypes in literary works and reveals the potential of LLMs to rectify gender stereotypes. This study raises awareness and promotes inclusivity within artistic expressions, making a significant contribution to the discourse on gender equality.
\\ \newline }

\begin{document}

\maketitleabstract

\blfootnote{* Equal contribution}

\section{Introduction}

\begin{figure}[t]
\begin{subfigure}{\columnwidth}
    \centering
    \includegraphics[width=0.75\textwidth]{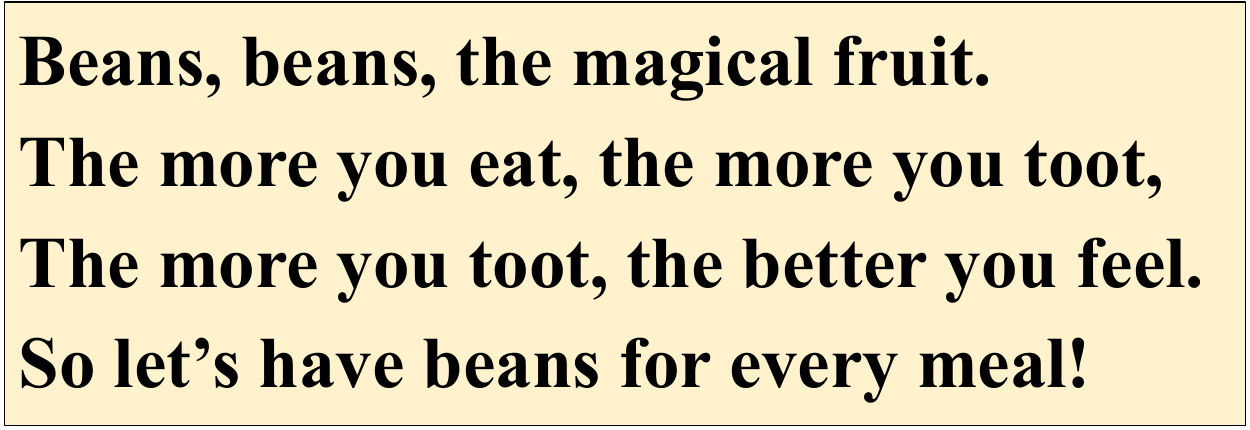}
    \caption{Non-stereotypical}
    \label{fig:fig1}
\end{subfigure}
\begin{subfigure}{\columnwidth}
    \centering
    \includegraphics[width=0.75\textwidth]{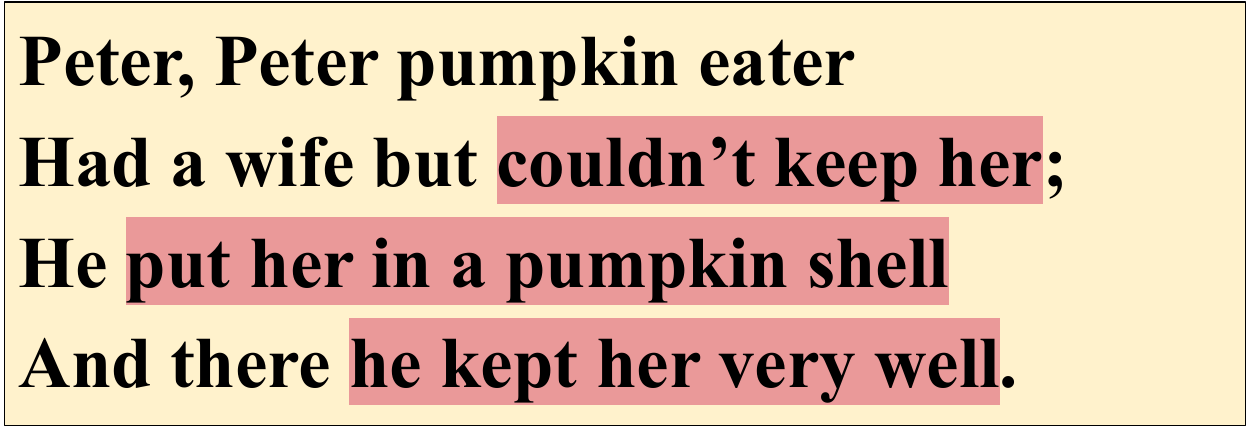}
    \caption{Stereotypical}
    \label{fig:fig2}
\end{subfigure}
        
\caption{Example of poems with and without explicit gender stereotype}
\label{fig:figs1_2}
\end{figure}

The acquisition of knowledge through experience is a fundamental process in the development of identity. Humans have the tendency to occupy roles pertaining to their gender \cite{blackstone2003gender}, which we call gender roles. Gender roles are socially constructed positions or behaviours that are learned and performed by individuals in accordance with their gender identity and the prevailing cultural norms \cite {bigler2007developmental}. This learning starts when toddlers are exposed to stories and rhymes, which are fundamental learning practices for language acquisition, and also facilitate their understanding of how society functions. Figure \ref{fig:fig1} is a rhyme that is taught to children, short and engaging, whilst communicating its inherent message clearly about the benefits of eating healthy foods like beans.

While many nursery rhymes convey valuable messages and teach language skills, some contain outdated messages that are no longer aligned with contemporary values. Figure \ref{fig:fig2}, a seemingly humorous poem about a troubled marriage, perpetuates patriarchal values by depicting the husband's control over his wife. Despite the progress that society has made, these rhymes continue to be taught to young children in nursery schools around the world.

By the time children reach adulthood, they internalize many gender stereotypes, whether consciously or unconsciously. These stereotypes can shape our thoughts, feelings, and behaviours in various ways. In the past, men have predominated over women in fields like politics, the military, and law enforcement, whereas in care-related professions like child care, health care, and social work, women have predominated over men \cite{sharma2016gender}. These professional roles are illustrations of typical male and female behaviour that has its roots in our culture \cite{diamond2002sex}. These findings lead us to define gender stereotypes in our study as generalizations or assumptions that are made about the typical characteristics, roles, and behaviours of men and women, based on their gender. These stereotypes are often internalized by individuals from a young age and can shape their beliefs, attitudes, and behaviours towards themselves and others \cite{Haines2016The}. Gender stereotypes can be harmful when they limit individuals' potential or create unfair expectations based solely on their gender, thereby promoting toxicity and misogyny \cite{Heilman2001Description}.

The predominant mode of knowledge acquisition stems from educational materials, such as textbooks and oral instruction. Notably, an inclination towards favouring males over females is observed within school textbooks, specifically across Southeast Asian nations \cite{islam2018gender}. The investigation reveals that the combined textual and pictorial indicators portray a significantly lower aggregate representation of females, accounting for merely 40.4 \%. Moreover, it is noteworthy that popular children's rhymes originating from the renowned collection known as Mother Goose perpetuate sexist ideologies and explicitly endorse gender stereotypes targeted at women \cite{nadesan1974mother}.

It could be argued that the rhymes and poems between the 18\textsuperscript{th} and 20\textsuperscript{th} centuries reflect the social norms of those eras. However, society has changed significantly since then, and some of the concepts that exist in these rhymes and poems are no longer considered appropriate. Therefore, educators need to be mindful of the stereotypes that may be present in these works and to critically evaluate their content before teaching them to students. This study tries to fill the gap by using various machine-learning techniques to reduce the amount of human intervention to rectify such stereotypes. The highlights of this study are threefold:

\begin{enumerate}
    \item \textbf{Dataset : }The identification of explicit gender stereotypes in rhymes and poems necessitates the availability of datasets. To the best of our knowledge, there are no publicly accessible repositories specifically dedicated to documenting poems and rhymes that contain explicit gender stereotypes. We created a publicly available dataset\footnote{\url{https://github.com/s-vigneshwaran/Revisiting-The-Classics}} of rhymes and poems collected from various sources, with each line manually annotated for gender stereotypes. This contribution will significantly advance research in this area, leading to a better understanding of the harms perpetuated by these rhymes and poems.

    \item \textbf{Classification : }Different AI models were trained and tested to classify whether a poem or a rhyme contains an explicit gender stereotype. Additionally, we propose a heuristic encoder, that utilizes annotator-learned features and sentiment analysis. The best model achieved an accuracy of 97\% while maintaining the recall at 0.81, signifying a better overall performance given that the dataset is highly imbalanced.

    \item \textbf{Rectification : } We test the efficacy of Large Language Models (LLMs) in rectifying gender stereotypes in rhymes and poems by conducting a survey that compares human and LLM rectifications. 
\end{enumerate}

\section{Related Work}\label{sec:related_work}

Developmental research has emphasized the importance of understanding and controlling stereotyping and prejudice in early childhood \cite{bigler2007developmental}. \citet{kane1996conceptualization}'s cross-cultural study revealed that children as young as two or three years old can label gender and classify objects accordingly, while by age four, they conform to societal gender norms. \citet{Fast2016ShirtlessAD} discovered that male over-representation and traditional gender stereotypes are commonly observed in online writing communities.

Early work using supervised tasks has demonstrated promising results in gender stereotype analysis, leveraging large amounts of unlabeled data to reduce error in gender classification \cite{Bergsma2009Glen}. \citet{hopark2018} developed methods to measure gender biases in language models trained on abusive language datasets, analyzing the impact of pre-trained word embeddings and model architectures. 

Word embeddings have been established to contain significant gender stereotypes, which correlate with real-world bias \cite{Bolukbasi2016Quantifying, Du2019Exploring}. \citet{Gonen2019How} followed suit by rectifying grammatical gender bias in word embeddings. Quantifying gender stereotypes in language corpora using word embeddings has revealed their consistent and robust presence, including theoretically selected stereotypes \cite{Charlesworth2021GenderSI}. However, traditional techniques for debiasing embeddings can worsen the downstream classifier's bias by providing a less noisy channel for communicating gender information \cite{Prost2019Debiasing}. 
Furthermore, experiments have shown an overall increase in the gender bias of neural models when they exploit transfer learning, particularly when using pre-trained embeddings that are already biased \cite{Rekabsaz2020Do}. \citet{lu2020gender} utilized counterfactual data augmentation to mitigate gender bias outperforming word embedding debiasing approaches.

In addition to gender bias, online toxicity related to gender identities, such as misogyny and sexism, has also been a major focus of research. Numerous studies have collected data from social media platforms on sexism and misogyny, which often contain linguistic complexities such as leet-speech and code-switching. Machine learning techniques have been successfully employed to classify this data \cite{rodriguez2020automatic, frenda2019online, chiril2020annotated}.

While these works have highlighted the importance of understanding gender-based threats, gender stereotypes and their presence in various language-related domains like social media platforms, there is a need to address gender biases in literary works, such as rhymes and poems, which are influential sources of cultural norms and societal roles. This study serves as a foundation for future work in this field to identify the presence of bias in children's rhymes and poems and strategies to identify gender stereotypes present in rhymes and poems.
We also propose effective strategies for mitigating the negative effects of gender-biased rhymes and poems on children by leveraging Large Language Models to rectify the rhymes and poems that align with modern-day values.

\section{Dataset}\label{sec:dataset}

The data collection process involved the acquisition of children's rhymes and adolescent-appropriate poetry from a variety of sources. These sources encompassed a broad range of content, including works by renowned poets such as Shakespeare and Frost, as well as popular collections such as Mother Goose \cite{GroverRichardson1915}. 

The selection process for the creation of a comprehensive dataset of children's rhymes and poems was designed to ensure diversity in terms of style, content, and cultural background. Specifically, in order to ensure that our dataset was diverse and accessible, we collected rhymes and poems from a variety of published sources, after extensive consultation with educators in the field of Literature and Education. This process ensured that the dataset was representative of a wide range of cultural backgrounds and perspectives and that the rhymes and poems were free from language errors and complexities. In addition to rhymes \& poems originally written in English, {we used 20 publicly available translated poems from 11 different languages (Refer to Table \ref{tab:language-stats} in Appendix)}. The goal is to create a rich resource for analysis and research that represents a wide range of rhymes and poems. The dataset used in this study contains a significant class imbalance. This could be attributed to the incomplete translation of cultural resources and the digitization of English resources. Additionally, not all of the English poems and rhymes that have been published have been made available online.

\

\noindent \textbf{Annotation} A two-phase annotation strategy was employed. The initial phase involves establishing annotation guidelines utilizing a subset of the dataset. The second phase employs the established guidelines to label the remaining dataset. A pair of annotators, aged 22 and 23, respectively, undertook the annotation of 50 rhymes and poems. Both annotators are non-native English speakers and have undergone 17 years of English language training, underscoring their qualifications to comprehend the intricacies and subtleties of the language. 

In the first phase, the annotators conducted an annotation procedure in which they were not aware of the identity of the poems or rhymes they were annotating. This was done to prevent any unconscious bias from influencing their annotations. In other words, the annotators were blind to the identity of the poems or rhymes, which means that they did not know which poems or rhymes contained explicit gender stereotypes. This was done to ensure that the annotators' annotations were as objective as possible. Regardless of the annotator's background, two annotators sufficed for the task of labelling rhymes and poems because this task is relatively linguistically simple. Rhymes and poems are taught to schoolchildren regardless of geographic location, so annotators from different backgrounds are likely to have similar knowledge of them.

The first phase continued for multiple iterations until they reached a satisfactory inter-annotator reliability score. Between iterations, annotators met to discuss and adjudicate any disagreements. The disagreements primarily revolved around the choice of words and the interpretation of the given lines. For instance, in lines such as \textit{"One for my \textbf{master}"} and \textit{"Wilt thou be \textbf{mine}?"}, particular attention was paid to the concept of ownership. It should be noted that the terms \textit{\textbf{master}} and \textit{\textbf{mine}} hold distinct connotations in terms of ownership, as \textit{\textbf{mine}} could be implied as the possession of the opposite gender, whereas \textit{\textbf{master}} does not connote ownership specific to a particular gender. Words relating to aestheticism like \textbf{\textit{pretty}} also had disagreements when their usage was tied to a particular gender but were decided to be non-stereotypical due to the subjective nature of beauty. In order to measure the inter-coder reliability for the annotation process, Krippendorff's $\alpha$ is used \cite{de2012calculating}. After four iterations of disagreement analysis, we attained Krippendorff's $\alpha$ of 0.96.

Phase two involved the establishment of guidelines by means of deliberating and evaluating the discussions that transpired between the iterations.

\begin{itemize}
    \item Lines containing words or phrases that could be interpreted as toxic or hateful towards a particular gender are classified as containing stereotypes.
    \item Stereotypes are identified in lines containing sexist words or phrases toward women.
    \item Lines are considered stereotypes if they contain gender-specific behaviour stereotypes.
    \item Sentences implying ownership over the other gender are classified as stereotypical.
    \item Stereotypical sentences include those suggesting that certain behaviour is exclusive to one gender.
    \item Sentences with terms that were historically honorific but now have negative meanings are classified as stereotypical.
\end{itemize}

\begin{table}[htbp]
\centering
\resizebox{\columnwidth}{!}{
\begin{tabular}{lrr}
\hline
\textbf{Label}                  & \multicolumn{1}{l}{\textbf{\# Verses}} & \multicolumn{1}{l}{\textbf{\# Lines}} \\ \hline
Stereotypical Rhymes            & 65                                    & 151                                    \\
Non-Stereotypical Rhymes        & 274                                   & 5,157                                  \\
Stereotypical Poems             & 80                                    & 359                                    \\
Non-Stereotypical Poems         & 242                                   & 7,647                                  \\
Augmented Stereotypes           & 290                                   & 1,347                                  \\ \hline
Total (Without Augmentation)    & 661                                   & 13,314                                 \\ 
Total (With Augmentation)       & 951                                   & 14,661                                 \\ \hline
\end{tabular}}
\caption{Dataset Statistic}
\label{tab:table1}
\end{table}

With the help of the guidelines, one of the two annotators labelled the rest of the remaining data. Table \ref{tab:table1} shows the number of poems and rhymes that are stereotypical in terms of lines and their entirety.

\

\noindent \textbf{Text Augmentation} Due to the class imbalance present in the dataset, as seen in Table \ref{tab:table1}, we used augmentation to increase the number of non-stereotypical poems in the training set. Data augmentation was performed by synthesizing synonym versions of the poems and rhymes in the training set using GPT-3.5 \cite{openai2021chatgpt}. The following prompt was used: "\textit{Replace} [*nouns or subject/objects from the poem or rhyme*]\textit{ with synonyms. Keep the poem rhyme scheme and sentence formation intact forcefully}". We designed this prompt specifically targeting nouns and synonyms in order to augment the text without affecting the bias present. The augmentation resulted in doubling the number of stereotypical rhymes and poems increasing the total number of stereotypical poems and rhymes to 290, as shown in Table \ref{tab:table1}. Table \ref{tab:table2} contains examples of original and augmented versions.

\begin{table}[ht]
\begin{tabular}{p{0.43\linewidth} p{0.43\linewidth}}
\hline
\textbf{Initial Sentence} & \textbf{Augmented Version} \\ \hline 
Jack and Jill went up the hill, To fetch a pail of water. & Jack and Jill went up the mountain, To obtain a bucket of water. \\
And when I bake, I'll give you a cake, & And when I fry, I'll give you a pie, \\ 
\hline
\end{tabular}
\caption{Examples of augmented text}
\label{tab:table2}
\end{table}

\section{Methodology}\label{sec:Methodology}

This section elucidates the methodologies employed for the detection of gender stereotypes through the utilization of multiple models. Furthermore, it delves into the functioning and mechanics of the proposed heuristic encoder.

\subsection{Heuristic Encoder} \label{subsec: HE}

\begin{figure*}[htbp]
    \centering
    \includegraphics[width=\textwidth]{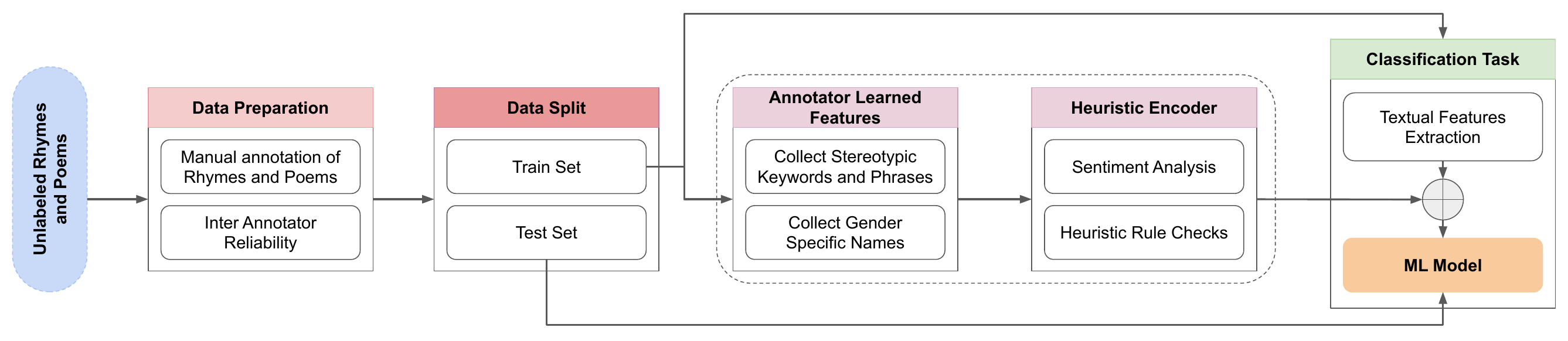}
    \captionsetup{justification=centering}
    \caption{The Heuristic Encoder: Incorporating annotator-learned features and sentiment analysis for gender nuances and sentiment orientations}
    \label{fig:fig3}
\end{figure*}

Generally, a model for classification uses only the ground truth for optimizing the predictions. However, studies have shown that knowledge-infused models perform better for specific tasks \cite{logan-etal-2019-baracks}. Inspired by this finding, we propose the concept of a Heuristic Encoder. The core idea behind the Heuristic Encoder involves the usage of annotator-learned features. In Heuristic Encoder, instead of using an external knowledge base, we use annotator-learned features to complement the input features to enhance the model’s prediction ability. To achieve this, the annotators were asked to compile a comprehensive list of words, phrases and gender-specific names that they consider stereotypical from the list of poems and rhymes that were annotated, making the feature list limited to the dataset for better contextual understanding. This list acts as an additional source other than the labels of the text. In addition, we also perform sentiment analysis as part of the encoding process, for which we employed off-the-shelf TextBlob \cite{loria2018textblob}. This enables the identification and characterization of sentiment patterns and tendencies within the text, contributing to a deeper understanding of the emotional aspects conveyed.

A binary valued feature vector is generated for the given text, using the annotator learned features collected during the encoder design process as part of the annotation process. The structure of each vector is defined as follows: [Male Names, Female Names, Stereotypes, Negative, Positive]. This vector is designed to encompass relevant information related to gender-specific names, identified stereotypes, and sentiment polarity. For example, the line \textbf{“had a wife but couldn’t keep her”} has the word \textbf{wife} signifying that it has female representation. The phrase \textbf{couldn’t keep her} is a stereotype signifying female ownership. Finally, the entire line has a negative sentiment. As a result, the resultant vector is [0, 1, 1, 1, 0]. By incorporating these vectors into the analytical methodology, we propose a Heuristic Encoder, as shown in Figure \ref{fig:fig3}, enhancing the computational model's ability to capture and represent gender-related nuances and sentiment orientations within the textual data. In order to prevent overfitting, the Heuristic Encoder uses annotator-learned features that belong to the train set. Consequently, there are words and phrases as part of the test set that the Heuristic Encoder does not use for feature augmentations.

The incorporation of the Heuristic Encoder assumes a significant role in enhancing traditional Machine Learning model's predictive accuracy by introducing supplementary features derived from the knowledge acquired through the annotators' expertise.

\subsection{Classification Methods}

We conducted an empirical study to investigate the effectiveness of different supervised classification methods for poems and rhymes in the dataset. We used variations of XGBoost and BERT-based models.

\

\noindent \textbf{XGBoost}: We utilized XGBoost as a baseline indicator. It should be noted that variations of input features were employed, leading to three different models. The efficacy of the Heuristic Encoder has been evaluated in conjunction with this particular machine-learning model.

\

\noindent \textbf{BERT}: Transformer-based architectures, such as the popular BERT model, have
shown impressive performance in a variety of downstream NLP tasks. We utilize a bert-base model\footnote{\url{https://huggingface.co/bert-base-uncased}} for fine-tuning our objective of stereotype classification.

\

\noindent \textbf{BERT$_{SS}$}: StereoSet is a dataset designed to quantify stereotypical bias in language models. This dataset comprises 17,000 sentences that assess model preferences in relation to gender, race, religion, and profession \cite{nadeem2020stereoset}. A BERT model fine-tuned on this dataset, which we call in this paper as {BERT$_{SS}$}, is publicly available\footnote{\url{https://huggingface.co/henryscheible/bert-base-uncased_stereoset_finetuned}}. We further fine-tuned this particular model since it aligns more closely with the requirements for our downstream stereotype classification task.

\

\section{Experiments and Results}
\label{sec:experiments}

\begin{table*}[htbp]
\centering
\begin{tabular}{lcccc}
\hline
\textbf{Model Name (nL/F)} & \textbf{Accuracy} & \textbf{Precision} & \textbf{Recall} & \textbf{F1-Score} \\ \hline
XGBoost (F) & 0.72 & 0.62 & 0.66 & 0.63 \\
XGBoost + HE (F) & 0.77 & 0.66 & 0.66 & 0.66 \\
XGBoost (2L) & 0.93 & 0.73 & 0.71 & 0.72 \\
XGBoost + HE (2L) & 0.94 & 0.74 & 0.73 & 0.74 \\
BERT Base (F) & 0.83 & 0.75 & 0.8 & 0.77 \\
BERT Base (2L) & 0.94 & 0.76 & 0.79 & 0.77 \\
BERT$_{SS}$ (F) & 0.82 & 0.74 & 0.69 & 0.71 \\ 
BERT$_{SS}$ (1L) & \textbf{0.97} & \textbf{0.77} & \textbf{0.81} & \textbf{0.79} \\
\hline
\end{tabular}
\captionsetup{justification=centering}
\caption{Model Performance Metrics[Macro Averaged]. \\Abbreviations: nL (number of lines), F  (full text), HE (Heuristic Encoder)}
\label{tab:table3}
\end{table*}

The experimental evaluation involved a comprehensive analysis of various models for the detection and classification of gender stereotypes. Table \ref{tab:table3} presents the performance metrics obtained from the different models.

\subsection{Experimental Setup} 

We analyzed our data using four different categorization schemes: monostichs (L), couplets (2L), tercets (3L), and full text (F). As our objective is to rectify stereotypical poems and rhymes, we consider a poem or rhyme to be stereotypical even if only one line contains a stereotype and a poem becomes a candidate for rectification if it is classified as stereotypical by the selected model.

After segmentation, the dataset is divided into a train set of 80\% and a test set of 20\%, but for BERT-based models, the train set was reduced to 70\% and the remaining 10\% is used for model validation. As a baseline, we utilized XGBoost with a maximum depth of 3 and 100 estimators. Regarding the XGBoost's input features, two distinct approaches were taken. The first approach utilizes the Word Frequencies of the vocabulary present in the dataset, and for the second approach, we concatenated the binary vector output of the proposed Heuristic Encoder (\ref{subsec: HE}) to the frequencies.

For BERT and BERT$_{SS}$ we fine-tuned the models with a learning rate of 2e-5 for 5 epochs using their respective tokenizers. The hyperparameters for training BERT-based models in our task are selected through repeated experimentation. A batch size of 16 was chosen, which strikes a good balance between training speed and hardware constraints. For the learning rate, we tried a variety of values (2e-5, 5e-5, 1e-7, and 4e-8) and numbers of epochs (4, 5, 15, 20). To avoid over-fitting, the number of epochs was manually fixed for each learning rate to observe the training process. Lower learning rates, such as 4e-8, were trained for longer periods (20 epochs) and learning rates such as 2e-5 and 5e-5 were trained for shorter periods (4-5 epochs). After experimenting with all these hyperparameter settings, it was observed that the learning rate of 2e-5 consistently performed best for all model variants.

In order to test the influence of contextual dependencies, the model's performance is tested with different input lengths like monostich (L), couplets (2L), tercets (3L) and the full text (F). It must be noted that the token lengths for the BERT-based models were changed according to the input lengths of the approach used.

\subsection{Results and Analysis}

As shown in Table \ref{tab:table3}, BERT$_{SS}$ (1L) is the best-performing model in terms of all the metrics. This intuitively makes sense since the model has an ingrained understanding of stereotypes and bias of a broader environment and here it adapts to the task of poems and rhymes classification. Notably, the BERT$_{SS}$ (1L) achieved a recall of \textbf{0.81} which is crucial as it signifies reduced false negatives, making it more important than precision in this context. The trend across the variations of models signifies that shorter text performs better when compared to longer texts. Models utilizing 3 lines (3L) of input have consistently under performed, therefore Table \ref{tab:table3} selectively presents the most promising models. Other model settings along with their results are presented in Table \ref{tab:appendix-1} (Appendix).

An interesting observation is how the proposed Heuristic Encoder is able to improve the model's performance by 5\% with longer text input, since the avenue for checking the heuristics is more and in shorter contexts, the addition of the Heuristic Encoder proves to improve important metrics like precision, recall, and F1-Score by 1-2\%. 

\section{Rectification}\label{sec:rectification}

Classical literature is a valuable resource for understanding the past, but it is important to consider the social and intellectual context in which it was produced. Compared to contemporary times, classic literature may contain outdated values regarding gender, race, class, and other social categories. \cite{prosic2015making} argues that many rhymes were created more than a hundred years ago when society cherished somewhat different values from those in the modern day. Care should be exercised when choosing the rhymes to be used in teaching modern-day children. By rewriting classic literature, writers can help to correct these biases and create a more accurate and inclusive representation of the past. It is important to note that rewriting classic literature is not about erasing the past. Rather, it is about re-imagining the past in a way that is more inclusive and representative of the diverse experiences of women and other marginalized people. 

\begin{table*}[t]
\begin{tabular}{p{0.3\linewidth} p{0.3\linewidth} p{0.3\linewidth}}
\hline
\textbf{Original Text} & \textbf{Human Rectification} & \textbf{ChatGPT Rectification} \\ \hline
Georgie Porgie, pudding and pie; Kissed the girls and made them cry & 
Georgie Porgie, pudding and pie; Kissed the girls and got into a fight.
& Georgie Porgie, friendly and kind; Shared a smile, left worries behind. \\
What are little girls made of? Sugar and spice, And everything nice, & 
What are little ones made of? Kindness and grace, A warm embrace,&
What are children made of? Kindness and courage, shining so bright, \\ \hline
\end{tabular}
\caption{Examples of Human vs ChatGPT rectification}
\label{tab:table4}
\end{table*}

\subsection{Rectification Process}

For the rectification process, an educator with over 20 years of experience in Montessori and primary education volunteered for the rectification process. Their daily teaching lessons, which prominently featured rhymes and poems, demonstrate their expertise in language development and engagement among young children, making them suitable for the task. From the dataset that we present as part of the paper, the educator picked 20 rhymes and 20 poems with varying literary devices and diversified content. These were then rectified to suit modern sentiments, while taking special care to alter the content with minimal changes to retain the aesthetics like rhyme scheme, sentence formation, etc. 

We also used a Large Language Model (LLM) to perform rectification since studies have shown LLMs to be effective language learning mediums, both as tutors and as independent learning tools \cite{Haristiani2019ArtificialI}. In this study, we use OpenAI's GPT-3.5 \cite{openai2021chatgpt} due to its established ability to pass various qualifying exams in law \cite{choi2023chatgpt}, medicine \cite{mbakwe2023chatgpt} and computer science \cite{bordt2023chatgpt} since its public release. For brevity, we will henceforth refer to GPT-3.5 as ChatGPT, since this is the only version used in our experiment.

To rectify the set of rhymes and poems, we employed the prompt: \textit{"Change the poem to remove gender stereotypes and make sure to keep sentence formation and rhyme scheme close to the original as much as possible."} An example of the original versus the ChatGPT rectified version is shown in Table \ref{tab:table4}.

\subsection{Validation}

A survey-based statistical analysis was undertaken to examine the differential rectification capacity of human participants compared to ChatGPT in adapting poems and rhymes to align with contemporary sentiments. From the rectified set from both human and ChatGPT, a specific subset of 5 rhymes and 5 poems incorporating gender stereotypes was selected, based on the discernment of the educator, with the intention of encompassing a diverse range of linguistic structures and content lengths to ensure variability. To ensure impartiality and eliminate potential bias, an evaluative survey was conducted in which the participants were unaware of the identity of the rectifiers (i.e., whether human or ChatGPT). The rectifications were randomly shuffled and presented to the participants as \textbf{Version 1} and \textbf{Version 2}, without any indication of which version was produced by which rectifier, while Table \ref{tab:table5} has used more descriptive labels \textbf{Human} and \textbf{AI} instead for better readability and convenience of the reader. This approach aimed to objectively gauge the effectiveness of human and ChatGPT rectification methods in mitigating gender stereotypes within rhymes and poems. 

\begin{table*}[h]
\centering
\begin{tabular}{p{0.75\linewidth} p{0.2\linewidth}}
\hline
\textbf{Question} & \textbf{Options} \\ \hline
\textbf{Q1.} Rate the level of gender stereotype reduction in {[}VERSION{]} & {[}1 - 5{]} \\
\textbf{Q2.} Rate the level of creativity in avoiding stereotype reduction in {[}VERSION{]} & {[}1 - 5{]} \\
\textbf{Q3.} Which version was successful in reducing gender stereotypes? & Human, ChatGPT, No Difference \\
\textbf{Q4.} Which version do you prefer in retaining the originality? & Human, ChatGPT, No Difference \\ \hline
\end{tabular}
\caption{Survey Questionnaire}
\label{tab:table5}
\end{table*}

\

\noindent \textbf{Survey Setting:} A total of 17 participants volunteered to answer the questions designed to assess the reduction of gender stereotypes, maintaining the originality and creativity of the rectified content as shown in Table \ref{tab:table5}. \textbf{Originality} in our survey setting refers to the degree of similarity between the rectified poem or rhyme and the original. \textbf{Reduction} refers to the extent to which gender stereotypes have been mitigated as part of the rectification process. This allows us to compare the two rectification mediums i.e. humans and ChatGPT. All 17 participants have completed at least 12 years of English training. Based on the survey data, we observed an age range of 18 to 63 years, with a median of 33. Among the 17 participants, 9 identified as females, 7 identified as males, and 1 chose not to disclose their gender identity.

\

\noindent \textbf{Survey Result} Based on the questionnaire, we formulated our hypothesis. \\

\noindent $H_0$: There is no significant difference between the two versions\\
$H_1$: There is a significant difference between the two versions.\\  

Due to the small participant sample pool, paired t-tests were conducted on two sets of data: one for reduction and another for creativity. To assess the level of gender stereotype reduction and creativity in the rectified version, participants were asked to rate the level of gender stereotype reduction on a scale of 1 to 5 and to describe how creative they perceived the version to be in avoiding gender stereotypes, given the context of the question.

\begin{table}[htbp]
\centering
\begin{tabular}{llll}
\hline
                    & {$t$-statistic} & {$p$-value} \\ \hline
\textbf{Reduction}  &  1.346            & 0.179     \\
\textbf{Creativity} & -0.809            & 0.419     \\ \hline
\end{tabular}
\caption{Statistical comparison of Reduction and Creativity: $t$-statistics and $p$-values}
\label{tab:table6}
\end{table}

\begin{figure}[H]
  \centering
  \includegraphics[width=\columnwidth]{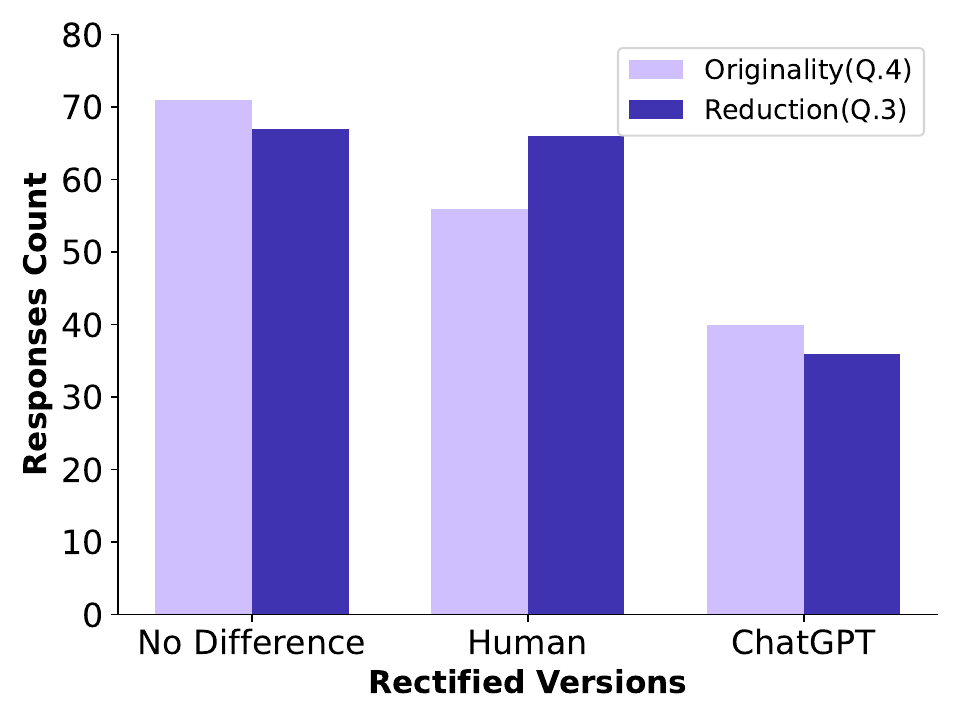}
  \caption{Gender Stereotypes: Originality and Reduction}
  \label{fig:fig4}
\end{figure}

Upon testing the hypothesis to compare the difference in gender stereotype reduction between human and ChatGPT rectification processes, it was determined that the p-value, exceeding the significance level of 0.05, led to the failure of rejecting $H_0$ suggesting a lack of evidence against the null hypothesis. Further studies with a larger sample size might be needed to detect a potential difference between the methods.
Detailed $t$-test results and corresponding $p$-values are presented in Table \ref{tab:table6}. Figure \ref{fig:fig4}, shows the response counts from questions 3 and 4 from Table \ref{tab:table5}.
The data, from Figure \ref{fig:fig4}, shows that the majority of participants found no difference in originality (Q.4) between human-rectified and ChatGPT-rectified text, while only a small number believed that human rectification was more original. Similarly, while the highest number of participants felt that humans were effective in reducing gender stereotypes (Q.3), an almost equal number of people believed there was no difference between humans and ChatGPT. This shows that ChatGPT is improving its capacity to correct rhymes and poems as well as humans.

The findings of this section could further bring about the following implications for research using LLMs for rectification of rhymes and poems:\\
\begin{enumerate}
    \item LLMs could be an effective tool in reducing gender stereotypes present in children's rhymes and poems without human intervention.
    \item LLMs may also be used to generate creative rhymes and poems that are free from gender bias and suit modern sentiments.
\end{enumerate}

\section{Conclusion}\label{sec:conclusion}

Gender stereotypes are pervasive in rhymes and poems, perpetuating biased perceptions and limiting the scope of individuals' identities. In this study, we investigated the presence of gender stereotypes in a diverse set of rhymes and poems from a variety of sources, creating an annotated dataset that will be a valuable resource for future research and addressing gender bias in classical literature. Explicit gender stereotypes were detected through the application of traditional machine learning models as well as BERT-based models. Furthermore, gender stereotypes were rectified using large language models (LLM) and human educators. A survey-based analysis compared the effectiveness of large language models (LLMs) and human educators in reducing gender stereotypes, and the results did not provide evidence of a substantial difference in their ability to mitigate gender bias. The findings of this study highlight the pervasive nature of gender stereotypes in literary works and also reveal the potential of LLMs in rectifying gender stereotypes.  Due to the limited availability of other large language models (LLMs) established in the field of education, GPT-3.5 was selected as the rectification medium for this experiment. Google Bard was still in the experimental stage at the time of writing, and LLaMA-2 was not yet available. However, we plan to explore the other LLMs (Google Gemini, Meta LLaMA-2) in the next installment of this work. By raising awareness and promoting inclusivity in artistic expressions, this research contributes to the discourse on gender equality.

\section*{Limitations}

The dataset contains an extensive collection of rhymes, but the selection of poems remains limited. Secondly, our annotation task involved only two annotators due to the low complexity of the task. However, as we expand this task to encompass multiple languages, it will be essential to involve additional annotators with varying levels of linguistic expertise to ensure accurate and comprehensive annotations. 
Our current rectification method is limited to the English language, but our findings lay the foundation for future work to overcome these constraints and enhance the scope and validity of our findings. The current sample size may limit our ability to detect subtle differences between human and ChatGPT rectification. Future studies with a larger sample size could potentially reveal more nuanced effects.

\section*{Ethics Statement}

This study does not involve any personal or public data pointing to an individual or a group of individuals thereby not breaking any ethical guidelines.

\section*{Acknowledgements}

We would like to thank Usha Sankaran for rectifying the poems and providing the rectification guidelines. Rajesh Sharma is supported by EU H2020 program under the SoBigData++ project (grant agreement No. 871042), CHIST-ERA grant No. CHIST-ERA-19-XAI-010 (ETAg grant No. SLTAT21096), and partially funded by CHIST-ERA project HAMISON.

\section*{References}
\bibliographystyle{template/lrec-coling2024-natbib}
\bibliography{references}

\appendix

\newpage
\onecolumn
\section*{Appendix} \label{appendix}

\begin{table}[!ht]
\centering
\begin{tabular}{lc}
\hline
\textbf{Language} & \textbf{Poems/Rhymes} \\ \hline
Italian & 3 \\
Portuguese & 3 \\
Spanish & 3 \\
French & 2 \\
German & 2 \\
Mandarin & 2 \\
Hindi & 1 \\
Japanese & 1 \\
Korean & 1 \\
Malay & 1 \\
Russian & 1 \\ \hline
\end{tabular}
\captionsetup{justification=centering}
\caption{Distribution of publicly available poems/rhymes translated from other languages}
\label{tab:language-stats}
\end{table}

\begin{table*}[hbp]
\centering
\begin{tabular}{lcccc}
\hline
\textbf{Model Name (nL/F)} & \textbf{Accuracy} & \textbf{Precision} & \textbf{Recall} & \textbf{F1-Score} \\ \hline

XGBoost (L) & 0.96 & 0.72 & 0.71 & 0.72\\
XGBoost (L) + HE & 0.95 & 0.7 & 0.7 & 0.7\\
XGBoost (2L) & 0.93 & 0.73 & 0.71 & 0.72\\
XGBoost (2L) + HE & 0.93 & 0.73 & 0.74 & 0.73\\
XGBoost (3L) & 0.91 & 0.66 & 0.65 & 0.66\\
XGBoost (3L) + HE & 0.91 & 0.67 & 0.66 & 0.66\\
XGBoost (F) & 0.72 & 0.62 & 0.66 & 0.63\\
XGBoost (F) + HE & 0.8 & 0.71 & 0.73 & 0.72 \\

& \multicolumn{1}{l}{} & \multicolumn{1}{l}{} & \multicolumn{1}{l}{} & \multicolumn{1}{l}{} \\
BERT (L) & 0.96 & 0.72  &   0.78    &   0.74 \\
BERT$_{SS}$ (1L) & \textbf{0.97} & \textbf{0.77} & \textbf{0.81} & \textbf{0.79}\\
BERT (2L) & 0.94 & 0.76 & 0.79 & 0.77 \\
BERT$_{SS}$ (2L) & 0.94 & 0.75 & 0.79 & 0.77 \\
BERT (3L) & 0.93 & 0.75 & 0.77 & 0.76 \\
BERT$_{SS}$ (3L) & 0.92 & 0.72 & 0.78 & 0.75 \\
BERT (F) & 0.83 & 0.75 & 0.8 & 0.77 \\
BERT$_{SS}$ (F) & 0.82 & 0.74 & 0.69 & 0.71 \\ \hline
\end{tabular}
\captionsetup{justification=centering}
\caption{Model Performance metrics of all task settings [Macro Averaged]. \\Abbreviations: nL (number of lines), F  (full text), HE (Heuristic Encoder)}
\label{tab:appendix-1}
\end{table*}

\end{document}